\documentclass[12pt]{article}

\usepackage{amssymb}
\usepackage{amsfonts}
\usepackage{graphicx}

\begin{document}

\title{Generalization in Learning: Eyring Formula and Predator-prey Model}

\author{S.~V.~Kozyrev\footnote{kozyrev@mi-ras.ru, Steklov Mathematical Institute of Russian Academy of Sciences, Moscow, Russia}, 
I.~A.~Lopatin\footnote{ilya.lopatin2606@gmail.com, Steklov Mathematical Institute of Russian Academy of Sciences, Moscow, Russia}, 
and A.~N.~Pechen\footnote{apechen@gmail.com, Steklov Mathematical Institute of Russian Academy of Sciences, Moscow, Russia, and Ivannikov Institute for System
Programming of the Russian Academy of Sciences,  Moscow, Russia}}

\maketitle

\begin{abstract}  
In  this work, generalization in machine learning is investigated
using a predator-prey model. We overview biological and chemical
interpretation of the stochastic gradient Langevin dynamics and
generative adversarial network. For a particular example of
recovering an unknown function by a polynomial of fixed degree from
a set of noisy data, the predator-prey model provides better
approximation compared to the gradient descent method. Further,
thermodynamical arguments, in particular Eyring formula, are used to
explain grokking (delayed generalization) phenomenon in machine
learning.
\end{abstract}


\section{Introduction}

In this work, we provide, following the general physical and
biological approaches to overfitting control in machine learning
developed in~\cite{KLP_Entropy}, a more detailed study of the
abilities of the predator-prey model to control overfitting using the
example of reconstruction of an unknown function from noisy data,
and outline possible application of thermodynamical arguments to
grokking. The standard approach to machine learning is based on the
optimization by the stochastic gradient descent \cite{Gasnikov}.
In~\cite{KLP_Entropy} and here, following~\cite{Parisi,Parisi1}, we
consider stochastic gradient descent in the form of the stochastic
gradient Langevin dynamics with random walk in the potential which
is given by the risk functional. This approach allows to apply known
methods of thermodynamics theory to machine learning. In particular,
the Eyring formula in kinetic theory~\cite{Eyring_1935} describes
selection of wide minima in optimization by stochastic gradient
Langevin dynamics and therefore explains better generalization
ability for stochastic gradient optimization. Here we consider
overfitting control in the framework of the algorithmic stability
approach, more precisely, by suppressing narrow minima of the
empirical risk functional. We also consider the generative
adversarial network (GAN) model~\cite{GAN} as a mathematical
predator-prey model in biology, where the discriminator and
generator in GAN play the roles of the prey and predator,
respectively. This allows to explain better generalization in GAN
model by the predator pushing the prey out of narrow
minima~\cite{KLP_Entropy}.

In Section~\ref{Sec:Stochastic}, the stochastic gradient Langevin
dynamics is described by a stochastic differential equation (SDE).
The Eyring formula of chemical kinetics allows to explain the
ability of better generalization for stochastic gradient
optimization in the presence of noise~\cite{KLP_Entropy}.

In Section~\ref{Sec:Grokking}, we explain the grokking
phenomenon~\cite{Grokking1} of delayed generalization using ideas of
statistical physics. In particular fast memorization is described by
a Brownian motion in potential (i.e., stochastic Langevin gradient
descent) while delayed generalization is explained by a Brownian
motion with zero potential.

In Section~\ref{Sec:GAN}, the GAN model is  described by a system of
two stochastic differential equations which possess a biological
interpretation as interaction between a predator and a prey.
Learning for GANs by solving a system of ordinary differential
equations was considered in~\cite{Qin,Oseledets,Kolter}.

In Section~\ref{Sec:NR}, we show for a particular problem of
recovering an unknown function by a polynomial of fixed degree from
a set of noisy data, that the predator-prey provides better
approximation compared to the gradient descent method.

Section~\ref{Sec:Conclusions} (the Conclusion) summarizes the results.

\section{Stochastic gradient descent and generalization}\label{Sec:Stochastic}

\subsection{Stochastic gradient Langevin dynamics}

The statement of the learning problem is as follows. Let be given a
sample $\{z_l\}$, $l=1,\dots,L$, a loss function
$\mathcal{L}(z,x)\ge 0$, and a parameter $x$ called hypothesis (we
assume that the hypothesis space is $\mathbb{R}^m$). If
$\mathcal{L}(z,x)=0,$ then the loss is absent. Minimization of the
empirical risk is the following problem
\begin{equation}\label{min}
 f(\{z\},x) = \frac{1}{L}\sum\limits_{l=1}^L \mathcal{L}(z_l,x) \,\to\, \min\limits_{x}.
\end{equation}
If not important, we will omit the sample $\{z\}$ variable in the following.

Gradient descent algorithm for this problem (if the loss  function
is continuously differentiable) is given by the differential
equation $\frac{df}{dt}=-\nabla f(x)$, which is numerically described by
the iterative process with gradient step $\alpha_k$ starting from
some $x_0$,
\[
x_{k+1}=x_{k} - \alpha_k \nabla f(x_{k}), \quad k=0,1,2,\dots.
\]
We describe the stochastic gradient descent by the
stochastic gradient Langevin dynamics
$$
x_{k+1}=x_{k} + w_{k} - \alpha_k \nabla f(x_{k}),
$$
where $w_{k}$ are small independent Gaussian  random vectors. In the
continuous limit such evolution reduces to the stochastic
differential equation
\begin{equation}\label{stochastic_equation}
d\xi^i(t)=\sqrt{2\theta}dw^i(t)- \frac{\partial f(\xi(t))}{\partial x^i} dt,
\end{equation}
where $\xi(t)=\{\xi^i(t)\} $ is vector of Brownian motion
coordinates $\xi^i(t)$, $dw^i(t)$ is the Wiener stochastic
differential and $\theta>0$ is the temperature.

The stochastic gradient Langevin dynamics  was discussed in
\cite{Parisi, Parisi1, Geman, Welling}.

The Fokker--Planck equation for the Langevin  dynamics
(\ref{stochastic_equation}) is the diffusion equation for the
distribution function $u=u(x,t)$ in the potential $f=f(x)\in
C^2(\mathbb{R}^m)$,
\begin{equation}\label{diff}
{\partial u\over \partial t}=\theta\,\Delta u+ \nabla u \cdot \nabla f+ u \Delta f,
\end{equation}
where $x\in\mathbb{R}^m$. This diffusion  equation equivalently can
be rewritten as
$$
{\partial u\over \partial t}=\theta\,{\rm\bf div}\,\left[e^{-\beta f}\,{\rm\bf grad}\,
\left[u e^{\beta f}\right]\right],
$$
where $\beta=1/\theta$ is the inverse temperature.  The Gibbs
distribution, which is up to normalization proportional to
$e^{-\beta f}$, is a stationary solution of Eq.~(\ref{diff}).

\subsection{Generalization and Eyring formula}

The lack of the ability to generalize for a solution  of the
learning problem (i.e., to have low risk on the training sample and
high risk on the validation sample) is called overfitting.
Overfitting is related to memorization instead of generalization for
a given training sample. We exploit the approach of wide (flat)
minima to generalization \cite{Stab1} which is related to the
algorithmic stability~\cite{Stab2,Stab3,Stab4}.

It is known that adding noise to the gradient  descent procedure
improves the ability to generalization~\cite{Gasnikov}. We will
discuss this as follows. Eyring formula of kinetic theory states
that the reaction rate [which is defined as the rate of transition
between two potential wells due to diffusion of the form
(\ref{diff})] is proportional to the exponent
\begin{equation}\label{Eyring}
e^{-\beta(F_1-F_0)},
\end{equation}
where $F_1$ is the free energy of the transition  state (the saddle
between the two potential wells) and $F_0$ is the free energy of the
initial state of the reaction (the potential well from which the
transition starts). The free energy of a state is $F=E-\theta S$,
where $E$ is the energy and $S$ is the entropy of the state. In the
general case, free energy of a set $U$ (say a potential well) is
defined as
$$
e^{-\beta F(U)}=\int\limits_U e^{-\beta E(x)}dx.
$$

The Eyring formula (\ref{Eyring}) implies  that the transition rate
will be lower (the free energy of the initial state will be lower)
for potential wells with higher entropy (wider ones). Thus, a
stochastic gradient Langevin dynamics (\ref{stochastic_equation})
corresponds to a regime in which wider wells more effectively
capture the learning system, i.e., algorithmically stable solutions
to the learning problem should be most likely selected during the
stochastic gradient optimization.

\section{Grokking, thermodynamics and Eyring formula}\label{Sec:Grokking}

In \cite{Grokking1}, the phenomenon of grokking (delayed
generalization) in learning theory was discovered and discussed
in~\cite{Grokking2,Grokking3}. In \cite{Grokking1}, training of a
neural network (a transformer consisting of a single-layer decoder)
to perform the algorithm of modular arithmetic -- addition of
residues modulo a prime number (the prime number was chosen as
$p=97$) was studied, other operations in the corresponding number
field were also studied (multiplication, division, calculation of
quadratic and cubic polynomials), and operations in the group of
permutations of five elements were considered. For such problems the
maximum size of the training sample is limited (in particular, for
addition of two residues modulo $p$ the maximum sample size is
$p^2$). It is also significant that the considered learning tasks
had a solution in the form of a ''structure'', i.e., a simple
algorithm (for example, an algorithm for addition modulo a prime
number). The network was chosen to be overparameterized, i.e., the
number of parameters was large, exceeding the size of the training
sample.

The grokking effect was discovered, which includes two phenomena:

1) Training the neural network to achieve zero risk on  the training
sample (memorizing the training set) took about $10^3$ iterations of
stochastic gradient descent. In this case, for the control sample
total overfitting was observed. But after continuing running the
stochastic gradient descent, after about $10^6$ iterations the
effect of generalization (grokking) was achieved -- almost zero risk
was achieved on the control sample and overfitting disappeared.
Regularization also found to have a positive effect on grokking.

2) The dependence of training time on the sample size  was studied,
a sharp increase in training time with a decrease in the sample size
was found. In this case, the points representing the training time
before achieving grokking approximately lie on a straight line in
logarithmic coordinates. On one axis, the logarithm of the time to
achieve $99\%$ accuracy on the control sample, i.e., grokking, on
the other axis, the sample size (the ratio of the size of the
training sample to the size of the maximum possible sample for a
given training problem). Therefore, the grokking time exponentially
decreases with increasing sample.

Thus, in grokking, at first there was a relatively  fast
memorization of the training sample, and then a significantly slower
disappearance of overfitting occurs. Let us explain the phenomenon
of grokking from the point of view of thermodynamics and statistical
physics.

Overparameterized networks are characterized by  the presence of a
whole region of parameters with zero risk for (\ref{min}), in which
$f(\{z\},x)=0$ and local minima merge into a manifold. Thus, in the
region of zero risk we have $\mathcal{L}(z,x)= 0$, $\forall z\in
\{z\}$. In order for the zero-risk region to contain the desired
solution in the form of an algorithm (e.g., modular arithmetic), the
entropy of the zero-risk manifold in practical situations should be
high enough.

The first phase of grokking -- memorization, i.e., achieving zero
risk on the training sample, consists of achieving the zero-risk
manifold $f(\{z\},x)=0$. Stochastic gradient descent in this case
consists of diffusion with drift in the region of non-zero gradient
of the risk functional, i.e., the distance traveled in the parameter
space towards the zero-risk manifold is approximately proportional
to the training time $t$ (the number of steps of the stochastic
gradient).

The second phase of grokking is described  by moving the system
along the zero-risk manifold towards the region of this manifold
with high entropy. This process occurs according to the second law
of thermodynamics (increase in the entropy). This movement has the
form of a Brownian motion along the zero-risk manifold  with zero
gradient, hence the traveled distance is proportional to $\sqrt{t}$
(the root of the number of steps of the stochastic gradient). It is
worth noting that in \cite{Grokking1} the first phase of grokking
took about $10^3$ steps of the stochastic gradient, and the next one
about $10^6$ steps (i.e., square of the memorization time).

As the size of the training sample $\{z\}$ increases,  the zero-risk
manifold $f(\{z\},x)=0$ decreases (since additional conditions
$\mathcal{L}(z,x)= 0$ are imposed for additional $z$). At the same
time, a solution in the form of an algorithm (''structure'') exists
for any training sample. Thus, as the training sample grows, the
zero-risk manifold shrinks toward a region with high entropy
containing the desired solution in the form of an algorithm.

Suppose that the entropy of the zero-risk  region decreases linearly
with increasing sample size, i.e., the volume of such a region
decreases exponentially. It can be assumed that imposing each
additional condition $\mathcal{L}(z,x)= 0$ for a new $z$ removes
approximately the same percentage of the volume of the zero-risk
manifold as the sample size increases. Such behavior will occur, for
example, in the case where the zero-risk manifold is (initially) a
multidimensional cube $[0,L]^m$, and
imposing the condition $\mathcal{L}(z,x)= 0$ for some $z$ will mean
imposing the constraint $0\le x\le\varepsilon L$, $\varepsilon<1$ on one
of the cube coordinates. Then, by choosing $z\in\{z\}$ independently
at each step and imposing a constraint on the corresponding
coordinate, we obtain an exponential decrease in the volume of the
zero-risk region during the iteration of such a procedure.

Eyring's formula $e^{-\beta(F_1-F_0)}$ gives  an approximation of
the dependence of the reciprocal of the grokking time on the sample
size. Here $F_0$ is the free energy of the initial grokking region
(part of the zero-risk manifold where the Brownian motion of
grokking begins, i.e., we assume that entropy decreases and $F_0$
grows linearly with the growth of the training sample size) and
$F_1$ is the transition state, i.e., the final grokking region (the
neighborhood of the solution to the learning problem, this free
energy changes weakly with the growth of the sample), in both cases
only the entropic parts of the free energy are important. Recall
that the free energy is $F=E-\theta S$, i.e., with the growth of the
entropy of the initial state (i.e., the decrease in the sample) the
transition time increases exponentially, which is observed in the
experiment \cite{Grokking1}.

\section{The GAN Model and generalization}\label{Sec:GAN}

The Generative Adversarial Network (GAN) model is a minimax problem  \cite{GAN}
\begin{eqnarray}
\label{minimax}
&\min_{y} \max_{x} V(x,y);&\\
\label{Vxy}
&V(x,y)=\frac{1}{L}\sum\limits_{l=1}^L \log D(z_l,x)  + \int\limits_{Z} p_{\rm gen}(z,y) \log(1-D(z,x)) dz.&
\end{eqnarray}
Here $z=\{z_l\}$, $z_l\in Z$ is a sample;  $D(z,x)$  and $p_{\rm
gen}(z,y)$, called the discriminator and the generator, are
parametric families of probability distributions on the space $Z$,
$x\in X$ and $y\in Y$  (we assume $X$ and $Y$ to be real vector
spaces) are parameters (hypotheses) for the discriminator and the
generator, correspondingly. In \cite{GAN}, the generator was
considered as a parametric family of mappings from some auxiliary
space to $Z$, these mappings transferred the probability
distribution on the auxiliary space to $Z$.

The functional $V$ in Eqs.~(\ref{minimax}) and~(\ref{Vxy}) has the form
$$
V(x,y)=V_1(x)+V_2(\rho(x,y)).
$$
The first contribution $V_1(x)$ is the  $\log$-likelihood function.
The second contribution $V_2(\rho(x,y))$ behaves qualitatively as
$-\rho(x,y)^{-1}$, where
$$
\rho(x,y)={\rm KL}(D(x)|p_{\rm gen}(y)),
$$
and the Kullback--Leibler distance between  probability
distributions $p$ and $q$ is given by the formula
$$
{\rm KL}(p|q)=\int_Z p(z)\log\frac{p(z)}{q(z)}dz.
$$

The stochastic gradient Langevin dynamics optimization procedure
for the problem (\ref{minimax}), (\ref{Vxy}) reduces to the system
of SDEs for random walks $\xi=(\xi^i(t))$, $i=1,\dots,m$,
$\eta=(\eta^j(t))$, $j=1,\dots,n$ on the spaces of the parameters of
the discriminator and generator
\begin{eqnarray}
\label{discriminator}
d\xi^i(t)&=&\sqrt{2\theta}dw^i(t)+ dt \frac{\partial}{\partial x^i}V(\xi(t),\eta(t)),\\
\label{generator}
d\eta^j(t)&=&\sqrt{2\theta}dv^j(t)- dt \frac{\partial}{\partial y^j}V(\xi(t),\eta(t)).
\end{eqnarray}
Here $w^i(t)$ and $v^j(t)$ are Wiener processes on the  parameter
spaces of the discriminator and generator correspondingly;
$\frac{\partial}{\partial x^i}V$ and $\frac{\partial}{\partial
y^j}V$ are derivatives of $V$ with respect to the first and second
arguments, respectively.

In \cite{Qin,Oseledets,Kolter}, oscillations of the parameters for
the gradient descent optimization of the GAN model were discussed.
In particular, for one-dimensional parameters $x$ and $y$ and
functional $V=\omega xy$ the above system of equations (for zero
noise) reduces to
$$
\frac{dx}{dt}=\omega y,\quad
\frac{dy}{dt}=-\omega x,
$$
and the solution oscillates around the minimax
$$
x=A\sin \beta\omega (t-t_0),\quad  y= A \cos\beta\omega (t-t_0).
$$

Arguments on overfitting reduction for stochastic  gradient descent
are also applicable to the system (\ref{discriminator}) and
(\ref{generator}) of SDEs for GAN~\cite{KLP_Entropy}.

The presence of the generator will suppress  overfitting more
effectively. By  (\ref{discriminator}) discriminator will tend to
reach local maximum of $V_1(x)$ and by (\ref{generator}) generator
will follow discriminator. This may be interpreted as a biological
prey (discriminator) and predator (generator) model, where the
predator pushes the prey out of narrow maxima of $V_1(x)$ to reduce
overfitting.

\section{Reduction of overfitting for function recovery with~a~predator-prey model}\label{Sec:NR}

\begin{figure}[!h]
\centering
\includegraphics[width = \linewidth]{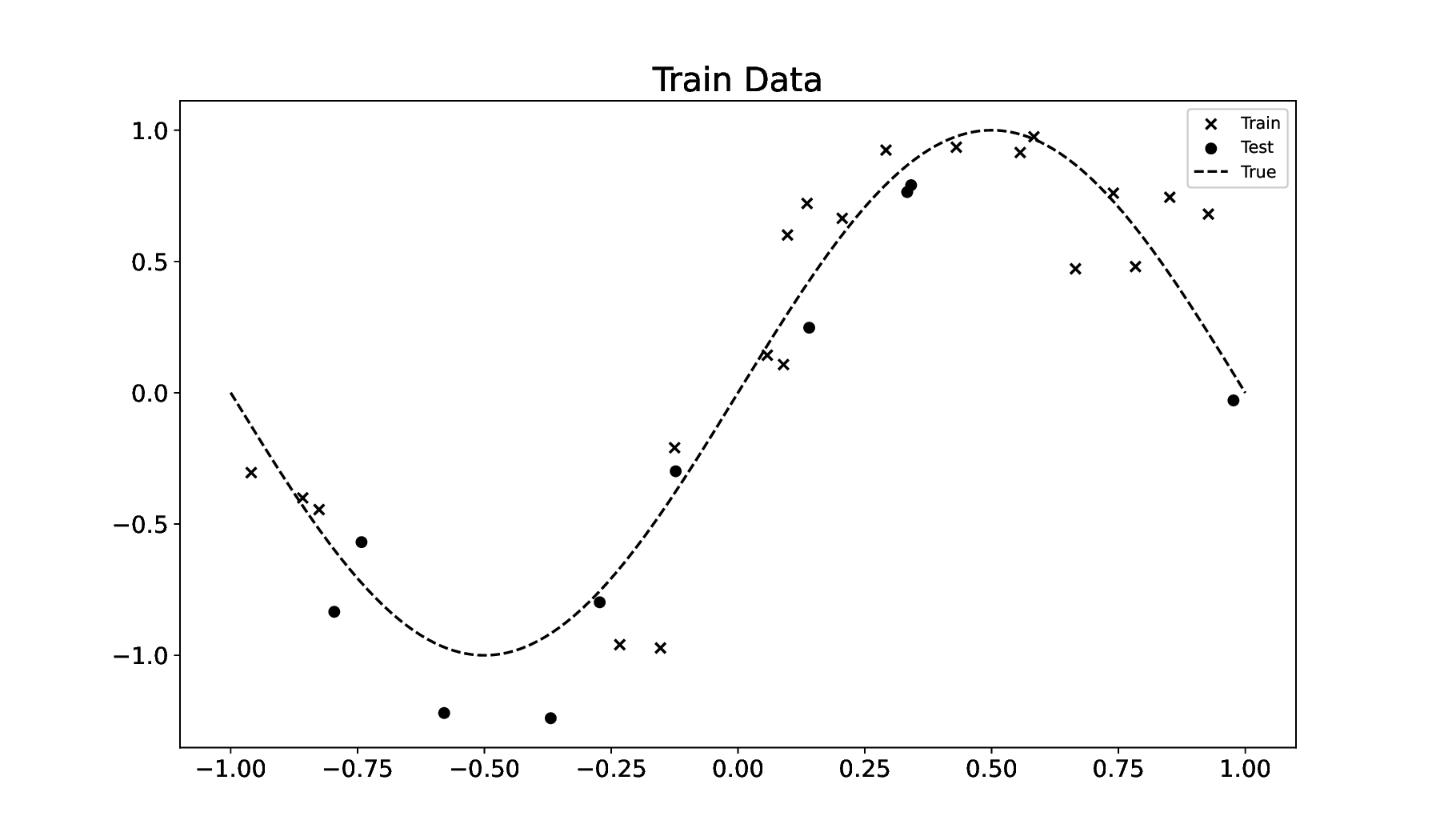}
\caption{Sampling examples of training and test datasets for the function $f(x)=\sin (\pi x)$.}
\label{Fig1}
\end{figure}

In this section, we numerically show the reduction of overfitting by
the ''prey-predator'' method introduced in~\cite{KLP_Entropy} for a
problem of reconstruction of an unknown function from a set of noisy
data (i.e., regression). Let $f$ be a hidden function, for which as
a particular example for simulations we consider
$$
f(x) = \sin (\pi x),\quad x \in [-1, 1].
$$
Let $\{ (x^{(1)}_{j}, y^{(1)}_{j}) \}_{j=1}^{N_1}$  be a training
dataset, where $y^{(1)}_{j} = f(x^{1}_{j}) + \xi^{(1)}_{j}$ with
$\xi^{(1)}_{j}$ being independent normally distributed random
variables with zero expectation and fixed dispersion. A test dataset
$\{ (x^{(2)}_{j}, y^{(2)}_{j}) \}_{j=1}^{N_2}$ is introduced
analogously. Sampling examples of the training and test datasets of
the hidden function $f=\sin (\pi x)$ with $N_1= 20$, $N_2=10$
produced by $\xi^{(1)}_{j}$, $\xi^{(2)}_{j}$ with standard deviation
$0.2$ are shown on Fig.~\ref{Fig1}.

We are looking for an approximation of the  hidden function $f$ by
polynomials of the degree $d$,
\begin{equation} \label{eq:Polynom}
g_{\theta} = \sum_{p=0}^{d} \theta_p x^p,
\end{equation}
which can be interpreted as a single-layer  neural network. To find
the coefficients $\theta_p$, we consider the problem of minimizing
the  mean squared error (MSE) loss on the training sample,
\begin{equation}
\mathcal{L} (\theta) =  \sum_{j=1}^{N_1} \left( y_j^{(1)} - g_{\theta} ( x_j^{(1)} ) \right)^2 \to \min .
\end{equation}

The quality of the model is determined by the  MSE on the test
sample.

\begin{figure}[!h]
\centering
\includegraphics[width =\linewidth]{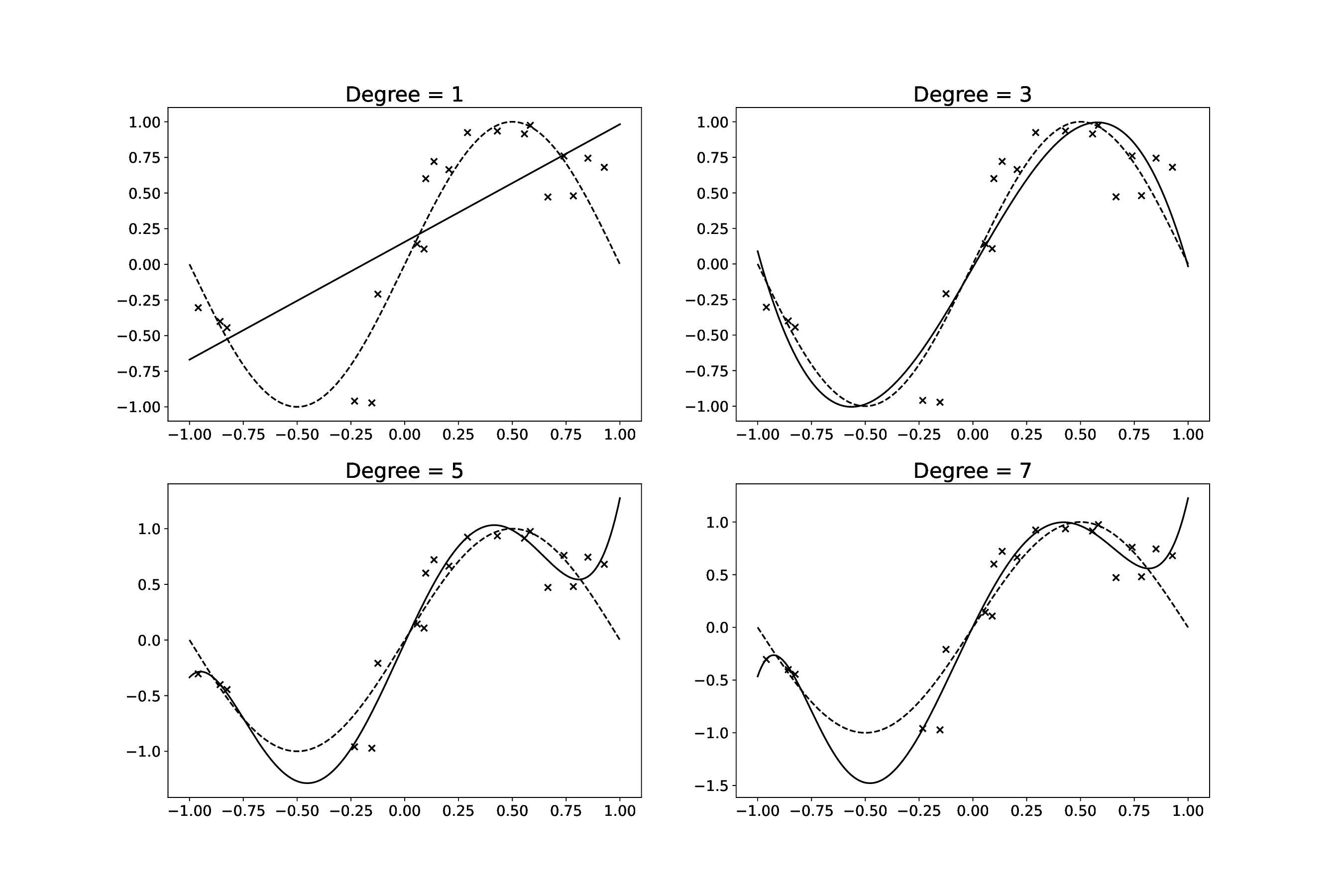}
\caption{Solid lines show approximations to the function $f$ (dashed lines) found by gradient descent after $10^6$ iterations for $d=1,3,5,7$ in (\ref{eq:Polynom}). The train points are marked by 'x'.}
\label{fig:OverfittingEffect}
\end{figure}

In the overfiting regime, loss on training  samples decreases while
loss on test samples increases. Let us consider the set of
approximating polynomials (\ref{eq:Polynom}) with degrees $d=1, 3,
5, 7$. For each degree, we run the same number of steps of gradient
descent with the same learning rate
\begin{equation}
\theta_{k} = \theta_{k-1} + \alpha \nabla_{\theta} \mathcal{L} (\theta_{k-1}).
\end{equation}
The results of the simulations are provided  in Table
\ref{tab:polynomial_results}  and Fig.~\ref{fig:OverfittingEffect}.

\begin{table}[h]
\centering
\begin{tabular}{|l|c|c|c|c|}
\hline
Degree $d$ & 1 & 3 & 5 & 7 \\
\hline
Loss on train & \, 0.171 \, &\,  0.072\,  &\,  0.027\,  & \, 0.026\,  \\
\hline
Loss on test & 0.400 & 0.033 & 0.161 & 0.159 \\
\hline
Deviation from $f$ & 0.227 & 0.007 & 0.059 & 0.080 \\
\hline
\end{tabular}
\caption{Learning results for different degrees of polynomials. The last line of the table shows the MSE of the model predictions for the hidden function on a uniform grid on $[-1, 1]$ with $1000$ points.}
\label{tab:polynomial_results}
\end{table}

The case $d=1$ corresponds to underfitting,  i.e., the model with
$d=1$ is unable to capture patterns in the data. The case $d=3$
corresponds to normal learning -- the model captures the patterns.
The cases $d = 5, 7$ correspond to overfitting, i.e., the model is
too complex and is affected by the noise in the data.

In \cite{KLP_Entropy}, we considered in learning a prey-predator
model given by a system of stochastic differential equations. The
prey $u$ and the predator $z$ with initial conditions $u_0, z_0$
were considered and their dynamics is given by
$$
u_k = u_{k-1} - \alpha_u \left( \nabla_{u} \mathcal{L}(u_{k-1}) - \overrightarrow{V}( u_{k-1} - z_{k-1} )  \right), 
$$
$$
z_k =  z_{k-1} + \alpha_z \frac{ u_{k-1} - z_{k-1} }{ \left| u_{k-1} - z_{k-1} \right| },
$$
where $\alpha_u$ and $ \alpha_z$ are constant learning rates,  the
term $\overrightarrow{V}$ provides repulsion of the prey from the
predator
$$
\overrightarrow{V}( \vec{p} ) = V( p ) \frac{\vec{p}}{p}.
$$
The following repulsion $V(p)$ was considered
$$
V(p) = \frac{A}{ 1 + e^{ c(p-l) } }  + C \frac{ e^{-\sigma p} }{p},
$$
where $A, C, c, l,$ and $ \sigma$ are the parameters of the
optimization. The first term is an approximation (at $c \to +
\infty$) of the step function with shift $l$, where $l$ is the
characteristic interaction distance, the second term is the Yukawa
potential providing the repulsion for short distances.

\begin{figure}[!h]
\includegraphics[width = \linewidth]{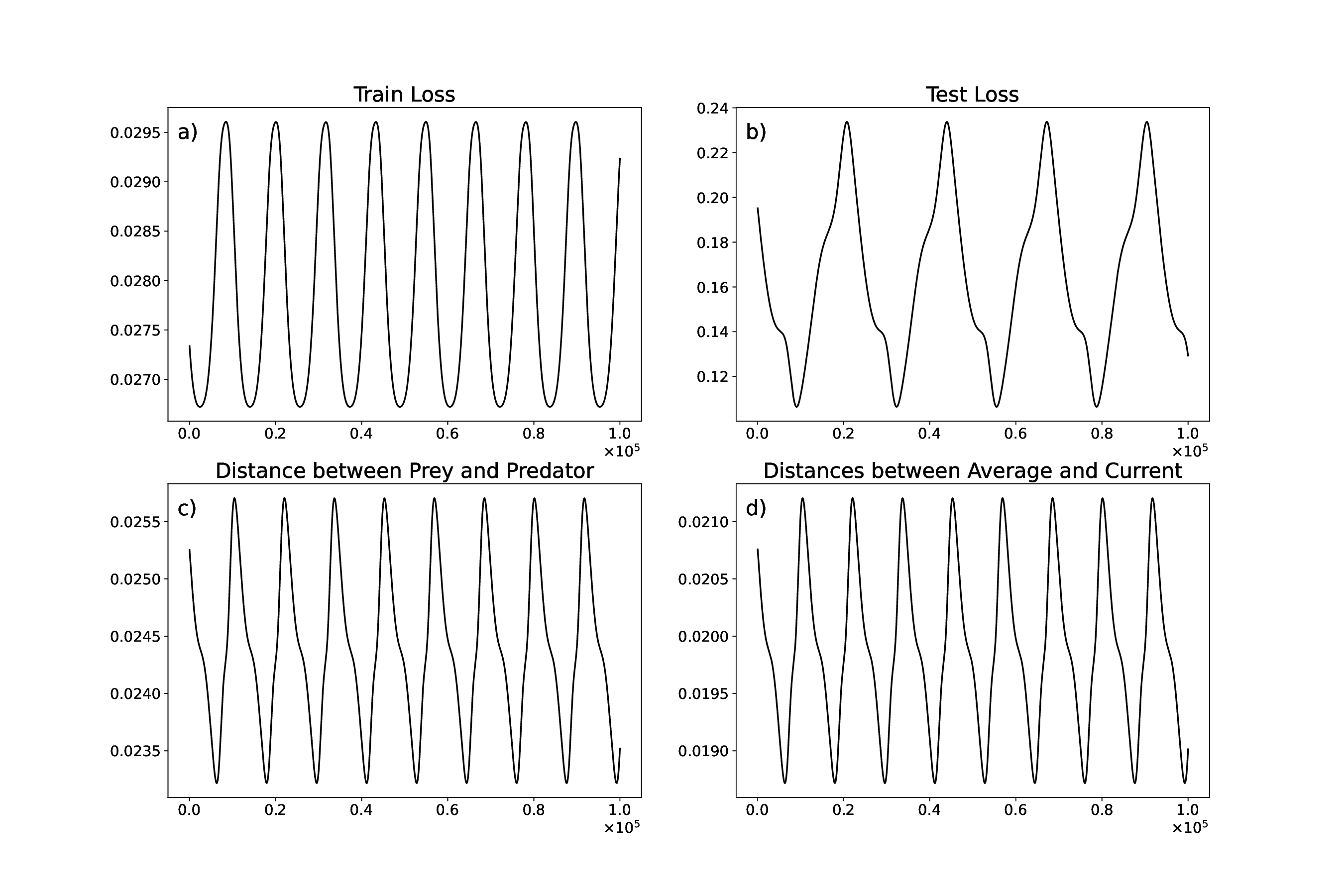}
\caption{Simulation results for: (a) loss on train; (b)  loss on
test; (c) distance between the predator and the prey; (d) distance
between the current state $u_k$ and the averaged state
$\overline{u}_k$ for the prey.} \label{fig:PreyPredatorProcess}
\end{figure}

On Fig.~\ref{fig:PreyPredatorProcess} the final $N=10^5$  of totally
$10^6$ iterations of the predator-prey model simulation are shown.

Here we use weighted moving average
$$
\overline{u}_k = \omega u_k + (1- \omega) \overline{u}_{k-1}.
$$
After that, it is possible to select the best model by the minimum
of the loss on test  for the last $N$ iterations for training by
ordinary gradient descent and predator--prey method. We obtain the
following results.
\begin{itemize}
\item The smallest MSE on the test sample at $N$ last iterations decreased from $0.161$ to $0.106$.
\item Deviation from the hidden function decreased from $0.0594$ to $0.0334$.
\end{itemize}

\begin{figure}[!h]
\centering
\includegraphics[width=0.7\linewidth]{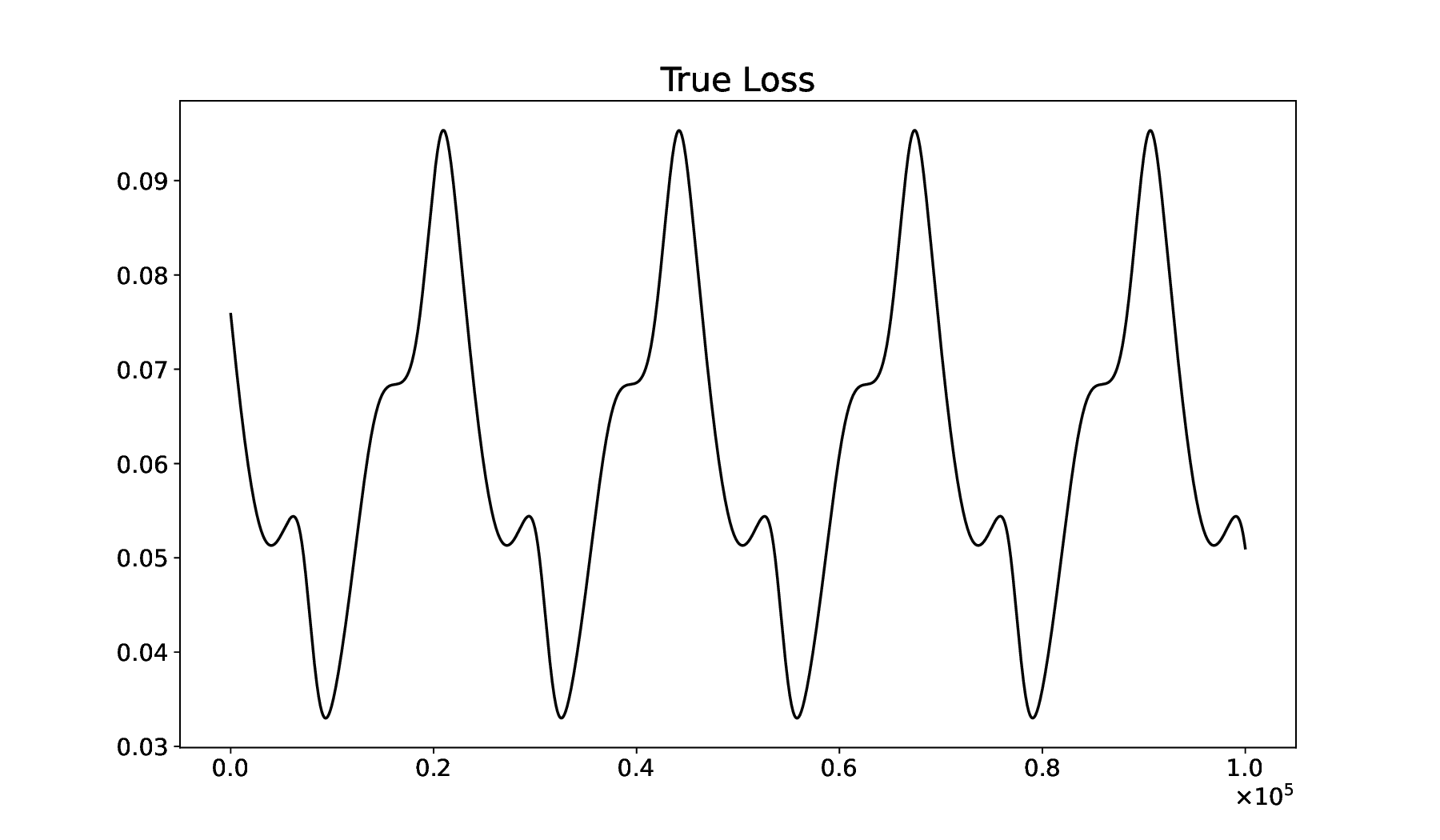}
\caption{Deviation of the model from the hidden function vs number of iterations.}
\label{fig:TrueLoss}
\end{figure}

Figure~\ref{fig:PreyPredatorProcess} shows the process of
oscillation corresponding to transitions  between several minima.
Moreover, as shown on Fig.~\ref{fig:TrueLoss} the deviation from the
hidden function behaves in a similar way.

A possible interpretation of this behavior is as follows. The graph
of the loss on train, besides extrema or ''wells'', also contains
wide ''lacunas'' containing several closely located minima. For
narrow minima instead of oscillations in one ''well'' we will
observe jumps between the neighboring minima. Different minima in
the same ''lacuna'' can have different loss on the test sample (up
to 1.5--2 times difference). The applied algorithm allows to search
several different nearby extrema and select the best ones. This kind
of behavior is observed for all runs of the ''predator--prey'' model
on other samples and with the same potential $V$.

In simulations it was observed that the trajectory  of the
(standard) gradient descent goes near the minimum of the loss
function. Similar behavior is observed for the predator--prey model
if the predator is sufficiently distant from the prey. The
experiments show a significant improvement only for a sufficiently
large number of iterations of the procedure.

Possible explanations for this effect are as  follows. In this
particular regression problem, we consider $\theta_0 = (0, 0, \dots,
0)$ as the initial $\theta$, which is close enough to the optimal
$\theta^*$ that can be obtained when training the model for $d=3$.
Therefore, it is possible that the model first converges to the data
''in general'' (i.e., the situation of normal fitting), and then,
with an oversize number of parameters, begins to be fitted to
specific, noise features of the sample which corresponds to
overfitting.

On the other hand, simulations for the ''predator--prey''  model
show the reduction of overfitting, this effect can reach up to
1.5--2 times reduction.

\section{Conclusion}\label{Sec:Conclusions}

In this paper, physical and biological approaches to explain better
generalization in machine learning are discussed. We use the
approach based on the selection of wide minima to control
overfitting. Then, description of stochastic gradient descent by the
stochastic Langevin dynamics (or Brownian motion in a potential)
explains selection of wide minima and improved generalization for
the stochastic gradient descent procedure. Moreover thermodynamic
methods allow to explain grokking which is delayed generalization
phenomenon. We also consider a biological-like predator-prey model,
where improved generalization is provided by the effect of pushing
out of prey from narrow maxima by the predator. We give a numerical
simulation for a simple learning problem which justifies this
statement.

\section*{FUNDING}
This research was funded by Ministry of Science and Higher Education
of the Russian Federation, grant number 075-15-2024-529.

\section*{ Conflict of interest }
The authors declare that they have no conflict of interest.

\end{document}